\documentclass[11pt]{article}
\usepackage[margin=1in]{geometry} 

\usepackage{graphicx} 
\usepackage{booktabs} 

\usepackage[toc,page]{appendix}

\usepackage{amsmath} 
\usepackage{amssymb} 
\usepackage{amsthm} 
\usepackage{mathtools} 
\usepackage{amssymb}
\usepackage[dvipsnames, svgnames, x11names]{xcolor}
\usepackage{algorithm}
\usepackage{algpseudocode}
\usepackage{times}

\usepackage{authblk}

\usepackage[bookmarks=true, colorlinks, breaklinks]{hyperref}
\AtBeginDocument{\hypersetup{
 linkcolor={NavyBlue},
 citecolor={NavyBlue},
 urlcolor={gray},
 }}
\usepackage{cleveref}

\usepackage[round]{natbib}

\usepackage[left]{lineno} %

\usepackage{mdframed}
\newmdenv[
  topline=false,
  bottomline=false,
  rightline=false,
  linecolor=gray,
  linewidth=1pt,
  skipabove=\topsep,
  skipbelow=\topsep
]{draftytext}

\title{\textbf{Agent-based imitation dynamics can yield efficiently compressed population-level vocabularies}}

\author[1]{Nathaniel Imel\thanks{Corresponding author: \texttt{n.imel@nyu.edu}}}
\author[2]{Richard Futrell}
\author[3]{Michael Franke}
\author[1]{Noga Zaslavsky}
\affil[1]{Department of Psychology, New York University}
\affil[2]{Department of Language Science, University of California, Irvine}
\affil[3]{Department of Linguistics, University of T{\"u}bingen}

\date{}

\begin{document}

\maketitle

\begin{abstract}
    Natural languages have been argued to evolve under pressure to efficiently compress meanings into words by optimizing the Information Bottleneck (IB) complexity-accuracy tradeoff. However, the underlying social dynamics that could drive the optimization of a language's vocabulary towards efficiency remain largely unknown. In parallel, evolutionary game theory has been invoked to explain the emergence of language from rudimentary agent-level dynamics, but it has not yet been tested whether such an approach can lead to efficient compression in the IB sense.  Here, we provide a unified model integrating evolutionary game theory with the IB framework and show how near-optimal compression can arise in a population through an independently motivated dynamic of imprecise strategy imitation in signaling games. We find that key parameters of the model---namely, those that regulate precision in these games, as well as players' tendency to confuse similar states---lead to constrained variation of the tradeoffs achieved by emergent vocabularies. Our results suggest that evolutionary game dynamics could potentially provide a mechanistic basis for the evolution of vocabularies with information-theoretically optimal and empirically attested properties.
\end{abstract}

\section{Introduction}

Although there are infinitely many possible languages, only a limited subset are actually spoken throughout the world. And while there are many conceivable meanings that words can have, actual word meanings are also constrained to a limited subset with considerable shared structure. Semantic typology aims to characterize this subset, asking which meanings are expressed across languages and why \citep{croftTypologyUniversals2003, kempSemanticTypologyEfficient2018}. 
To date, several robust semantic universals have been identified \citep{berlinBasicColorTerms1969, barwiseGeneralizedQuantifiersNatural1981, fintelUniversalsSemantics2008}, raising the question of their ultimate origin. One promising avenue of explanation, rooted in functionalist traditions, is that attested meaning systems reflect communicative pressures \citep{zipf1949least, roschPrinciplesCategorization1978, canchoLeastEffortOrigins2003, levinsonKinshipHumanThought2012, kempSemanticTypologyEfficient2018, gibsonHowEfficiencyShapes2019}. A common intuition for this idea is as follows: a language can be maximally simple, or it can be expressive, but it cannot be both maximally simple and maximally expressive simultaneously. Human languages might, however, evolve under pressure to optimally trade off simplicity and expressivity, and in doing so, substantially constrain the kinds of meanings that they lexicalize.

This broad notion of efficient communication has been formally grounded by \citet{zaslavskyEfficientCompressionColor2018} in a fundamental information-theoretic principle known as the Information Bottleneck~\citep[IB;][]{tishbyInformationBottlneckMethod1999}. IB has been derived from rate-distortion theory~\citep{shannonCodingTheoremsDiscrete1959, harremoesInformationBottleneckRevisited2007, zaslavskyInformationTheoreticPrinciplesEvolution2020}, the mathematical theory of lossy data compression. \citet{zaslavskyEfficientCompressionColor2018} argue that human languages evolve under pressure to optimally compress meanings into linguistic forms by satisfying the IB tradeoff between the informational complexity (or compressibility) and communicative accuracy of the vocabulary. This formal, quantitatively-testable prediction has been gaining broad empirical support based on extensive synchronic cross-linguistic data~\citep{zaslavskyEfficientCompressionColor2018,zaslavskySemanticCategoriesArtifacts2019,zaslavskyLetsTalkEfficiently2021,mollicaFormsMeaningsGrammatical2021,chenInformationtheoreticApproachTypology2023}, as well as direct diachronic evidence documenting language change~\citep{zaslavskyEvolutionColorNaming2022}. As noted by \citet{levinsonKinshipHumanThought2012}, however, little is known about the actual agent-based and cultural dynamics that could drive languages toward optimality. 

In parallel, another significant body of work has drawn on concepts from evolutionary game theory (EGT) to investigate how shared meaning systems can emerge from local interactions between agents~\citep{lewisConventionPhilosophicalStudy1969, steelsSelforganizingSpatialVocabulary1995, nowakEvolutionLanguage1999, pawlowitschWhyEvolutionDoes2008, skyrmsSignalsEvolutionLearning2010, hutteggerDynamicsSignalingGames2014, frankeVaguenessImpreciseImitation2018, spikeMinimalRequirementsEmergence2017}. This literature has primarily focused on the foundational question of how meaning can emerge \textit{ex nihilo}---how shared vocabularies can evolve in signaling games from exceedingly simple learning and evolutionary dynamics. From a broader perspective on language evolution, however, a significant conceptual gap remains: while EGT can explain how systems optimized for communicative success arise, it has not yet been explored whether such local success can yield the global information-theoretic efficiency that is empirically observed in human languages at the population level. Specifically, because EGT dynamics are typically driven by local utility rather than explicit optimization, it is not obvious that vocabularies that are evolutionarily stable in prominent EGT settings are also optimally efficient in the IB sense.

With regard to the first challenge---of identifying process-level explanations of efficiency---several studies have begun to investigate various evolutionary and learning dynamics. For example, \citet{xuHistoricalSemanticChaining2016} explored a process of semantic chaining as a potential mechanistic account of how communicative efficiency may emerge, though without considering agent-based dynamics. \citet{carstensenLanguageEvolutionLab2014} and \citet{carrSimplicityInformativenessSemantic2020} considered the dynamics of cultural transmission via iterated language learning in the lab, and \citet{kagebackReinforcementlearningApproachEfficient2020} explored reinforcement learning dynamics in artificial agents. While these studies provide valuable insight into possible mechanisms by which efficient communication systems may emerge, they each employ a different formulation of efficiency, thus lacking a consistent theoretical framework. In particular, they do not shed light on how languages might optimize the IB tradeoff. More recent work has started to explore the emergence of IB-optimal communication systems in multi-agent reinforcement learning settings \citep{chaabouniCommunicatingArtificialNeural2021,tuckerTradingUtilityInformativeness2022, tuckerHumanLikeEmergentCommunication2025, gualdoniBridging2024, carlssonCulturalEvolutionIterated2024}, yielding promising results as to the ability of interactive deep-learning agents to develop efficient human-like communication systems. These studies, however, heavily rely on model implementation decisions, such as the agents' underlying deep-learning architecture and hyper-parameters, and more fundamentally, they focus only on dyadic communication without addressing how such pair-level behaviors scale to and shape cultural evolution at the population level. 

Furthermore, regarding the second open question---whether EGT optimality leads to efficient compression in the IB sense---there is, to our knowledge, no prior research that establishes a link between EGT concepts of success in communication games and the IB framework.\footnote{Though see \citet{plotkin2000language} and \citet[Chapter 3]{skyrmsSignalsEvolutionLearning2010} for work connecting language evolution and information theory, and \citet{HarperMarc2009IGaE} for more connections between EGT and information theory more generally.} Therefore, the previous literature leaves open two key questions: (1) what are the core, cultural evolutionary mechanisms by which a population's language can become near-optimally efficient within the IB theoretical framework? (2) Are communication strategies that are optimal in EGT also information-theoretically optimal in the IB sense?

Here, we aim to address these two open questions by considering a model of the cultural evolution of population-level behavior and analyzing it with respect to the IB theoretical bound for vocabulary efficiency. We draw on the fundamental dynamical model of evolutionary game theory, the {replicator dynamic}~\citep{taylorEvolutionaryStableStrategies1978,Smith1982,hofbauerEvolutionaryGamesPopulation1998}, which describes how strategies or traits evolve based on their success, relative to the rest of the population. Specifically, we apply the social imitation dynamic proposed by \cite{frankeVaguenessImpreciseImitation2018} in the context of noisy, sim-max signaling games. Our approach does not require parameterizing agents with any particular learning algorithm, cognitive architecture, or priors, nor does it make strong assumptions about agents' rationality in these games. The result of this abstraction is an account of how myopic optimization grounded in local interactions between agents  can lead to the emergence of information-theoretically efficient population-level vocabularies.

The contributions of our work are twofold. First, we provide a unified theoretical perspective on efficient compression in a class of Lewis-Skyrms signaling games by formally integrating the IB framework of ~\citet{zaslavskyEfficientCompressionColor2018} with the noisy sim-max game and an accompanying population dynamic of ~\citet{frankeVaguenessImpreciseImitation2018}. Second, using model simulations, we show that (i) this dynamic can lead populations of agents to evolve IB-efficient vocabularies; (ii) contextual pragmatic standards in the game systematically constrain the specific tradeoffs of emergent systems; and (iii) noisy imitation of strategies limits the emergent systems' acheivable accuracy. Taken together, our work bridges two broad theoretical approaches to language evolution and demonstrates that local dynamics of imitation can drive population-level vocabularies towards information-theoretically optimal tradeoffs of the kind observed in human languages.

The remainder of the paper is structured as follows. In \Cref{sec:background}, we provide background on the IB framework for semantic systems (\Cref{sec:ib}), and an overview of \cite{frankeVaguenessImpreciseImitation2018}'s {noisy sim-max game} and accompanying population dynamic (\Cref{sec:simmax}). In \Cref{sec:unified-framework}, we introduce our unified framework that integrates efficient communication, as instantiated by IB, and evolutionary game theory, as instantiated by the imprecise imitation dynamic of the noisy sim-max game. We introduce a synthetic domain of idealized numerosity as our primary test case in \Cref{sec:test_case}, followed by the presentation of our main simulation results in \Cref{sec:results}. We discuss the implications of these results and conclude in \Cref{sec:discussion}.

\section{Background}
\label{sec:background}

\subsection{The Information Bottleneck framework for efficiently compressed semantic systems}
\label{sec:ib}

\cite{zaslavskyEfficientCompressionColor2018} argued that in order to support efficient communication, languages must compress meanings into words via a general information-theoretic principle known as the Information Bottleneck~\cite[IB;][]{tishbyInformationBottlneckMethod1999}. IB is an extension of rate distortion theory, the branch of information theory that focuses on optimizing lossy data compression given limited resources \citep{shannonCodingTheoremsDiscrete1959, bergerRateDistortionTheory1971, harremoesInformationBottleneckRevisited2007}. Applied to language, \citeauthor{zaslavskyEfficientCompressionColor2018} shows that the IB principle can be instantiated as a tradeoff between the informational complexity, or compressibility, and communicative accuracy, or informativeness, afforded by the vocabulary. We briefly summarize this formal framework below and in \Cref{fig:main_diagram_figure}A, keeping similar notation as \cite{zaslavskyEfficientCompressionColor2018}.

Consider a semantic category system equipped with an inventory of words $\mathcal{W}$ to communicate about a set of world states, $\mathcal{U}$. Let $u_t\in\mathcal{U}$ be a target world state that a speaker needs to communicate, drawn from a communicative need distribution $p(u_t)$. Since speakers may have uncertainty over the observed world state, the speaker's meanings are taken to be mental representations, or \textit{beliefs} over world states, defined by a probability distribution $m_t(u) \equiv p(u | u_t)$ that is associated with each target world state $u_t$. Therefore, $p(m_t)=p(u_t)$. A communication system is defined by a production distribution $q(w | m)$ of one (or a mixture) of the language's speakers, also referred to as an encoder, and by the interpretations a listener assigns to each word in the vocabulary. Assuming that the listener is Bayesian with respect to the speaker, each interpretation is defined as an estimator $\hat{m}_w(u) = \sum_{m} m(u) q(m | w)$, which corresponds to the listener's reconstruction of the speaker's belief state (or internal mental representation).


Optimal communication systems in this framework satisfy the IB complexity-accuracy tradeoff. In IB, both complexity and accuracy are defined in terms of mutual information, which is the most general measure of dependence between two random variables~\citep{coverElementsInformationTheory2006}. Given two (discrete) random variables $X$ and $Y$, the mutual information $I(X;Y)$ can be written as the Kullback-Leibler (KL) divergence between the joint distribution of the variables and the product of their marginals, 
\begin{align}
    \label{eq:mi}
    I(X;Y) =& D \left[p(X,Y) \| p(X)p(Y) \right]\\
           =&  \sum_{x,y} p(x,y) \log \frac{p(x,y)}{p(x)p(y)}.
\end{align}
The minimum of this quantity is achieved at $0$ when $X$ and $Y$ are statistically independent, i.e. $p(X,Y) = p(X)p(Y)$; as $X$ and $Y$ become more dependent, the information that $X$ and $Y$ carry about each other increases.

Given the discrete random variables for the speaker's meaning $M$ and word $W$, the {complexity} of a communication system is defined by the encoder's information rate:
\begin{equation} 
I_q(M;W) = \sum_{m,w} p(m) q(w | m) \log \frac{q(w | m)}{q(w)}.
\label{eq:complexity}
\end{equation}
The interpretation of this quantity roughly corresponds, on average, to the number of bits required to encode meanings into words (or any other communication signal). Maintaining a low information rate corresponds to using fewer bits for communication, and intuitively, to promote efficient communication, this quantity should be minimized to the extent possible. However, the complexity of the system trades off with its {accuracy}, defined as
\begin{equation}
I_q(W;U) = I(M; U) - \mathbb{E}_{q} \left[ D[M \| \hat{M}] \right].
\label{eq:accuracy}
\end{equation}
where the KL divergence $D[M\|\hat{M}]$ is the distortion between the speaker and listener meanings. 
Note that $I(M;U)$, the information that the speaker's meaning $M$ carries about the true world state $U$, does not depend on the encoder nor on the listener's interpretations. Therefore, maximizing accuracy in \Cref{eq:accuracy} amounts to minimizing distortion~\citep{harremoesInformationBottleneckRevisited2007,zaslavskyInformationTheoreticPrinciplesEvolution2020}.  Taken together, a language's semantic system is {efficient} in the IB sense to the extent that it succeeds at minimizing the overall IB objective function
\begin{equation}
  \label{eq:langrangian}
  \mathcal{F}_{\beta}[q] = I_q(M;W) - \beta I_q(W;U),
\end{equation}
\noindent where $\beta\ge1$ specifies the tradeoff between minimizing complexity and maximizing accuracy. The solutions to this optimization problem define the IB theoretical bound of efficiency (see \Cref{fig:compare_on_bound} for example in our context). It is mathematically impossible for a system to lie above this bound; however, Zaslavsky and colleagues have shown that across many domains \citep{zaslavskyEfficientCompressionColor2018,zaslavskySemanticCategoriesArtifacts2019,zaslavskyLetsTalkEfficiently2021, mollicaFormsMeaningsGrammatical2021, taliaferro2025bilinguals}, the encoders corresponding to natural language semantic systems achieve near-optimal compression, i.e., they lie very close to the IB bound. These findings naturally raise the following question: what cultural evolutionary mechanisms drive semantic systems towards efficiency? In this work, we address this question by turning to a prominent line of work within the evolution of language: the evolutionary dynamics of signaling games.

\subsection{Language evolution via the population dynamics of signaling games}
\label{sec:simmax}

The signaling games literature has long focused on how meaning can emerge \emph{ex nihilo} via simple communication games and minimalistic adaptive dynamics. The philosopher David Lewis introduced the signaling game to model conventional meaning in terms of strategic interaction  \citep{lewisConventionPhilosophicalStudy1969}, described as follows: A Sender observes a private state and sends a signal to a Receiver, who then takes an action, and both players receive payoff if the Receiver's action matches the Sender's intended state. In the most austere version of Lewis's game, there are two states, signals and actions. Over the decades, many researchers have explored many extensions of Lewis's original game, as well as various Nash equilibria and the behavior of various evolutionary and learning dynamics of signaling (and related) games \citep[i.a.]{crawfordStrategicInformationTransmission1982, steelsSelforganizingSpatialVocabulary1995, skyrmsSignalsEvolutionLearning2010, spikeMinimalRequirementsEmergence2017, Wechsler2025Emergence}. 

Here, we study a noisy variant of similarity-maximizing, or \emph{sim-max} \citep{jagerEvolutionConvexCategories2007, jagerVoronoiLanguagesEquilibria2011, oconnorEvolutionVagueness2014} signaling games, in which similar states of the world may be confused for each other. This game is played as follows: first, nature selects a state of the world $x_a \in \mathcal{X}$ with probability $p(x_a)$. A Sender $S$ observes a possibly different state $x_o \in \mathcal{X}$ with probability $p(x_o | x_a)$ due to the confusability of similar states. Sender then selects a signal $w \in \mathcal{W}$ with probability $S(w|x_o)$ to send to a Receiver, who can only observe the signal. Upon observing this signal, the Receiver intends to interpret $\hat{x}_o \in \mathcal{X}$. Due to confusability of similar states, Receiver $R$ actually realizes state $\hat{x}_a$ with probability $p(\hat{x}_a | \hat{x}_o)$ as the final reconstruction of $x_a$. Nature awards payoff to both players to the extent that the finally reconstructed state $\hat{x}_a$ is similar to the actual state $x_a$. The goal of both Sender and Receiver is to maximize the expected similarity between the Sender's intended state and the Receiver's reconstruction. 

There are two main reasons why this particular kind of signaling game is well-suited to study the evolution of efficiently compressed semantic systems. First, this game is independently motivated, originating in earlier work that modeled the evolution of vagueness in both synthetic domains \citep{frankeVaguenessImpreciseImitation2018, correiaEcologyVagueness2019} as well as the human color-naming domain \citep{correiaMoreRealisticModeling2019}. However, it has not yet been studied in the context of efficiency in semantic systems. Second, unlike classic Lewis-Skyrms games, which assume perfect perception and use binary payoffs, this noisy sim-max game incorporates perceptual uncertainty and graded similarity-based payoffs. This aligns nicely with the speaker's perceptual uncertainty captured by the IB framework, offering a minimal setting for studying the relationship between Lewis-Skyrms games and the IB efficiency principle. 

Following prior work on modeling the evolution of communicative behavior in signaling games \citep{skyrmsSignalsEvolutionLearning2010, pawlowitschWhyEvolutionDoes2008, hutteggerEvolutionaryDynamicsLewis2009, hutteggerDynamicsSignalingGames2014, nowakEvolutionLanguage1999, nowakEvolutionUniversalGrammar2001, frankeVaguenessImpreciseImitation2018, jacobConsensusGameLanguage2023}, we will model the evolution of communication based on the principles of frequency-dependent selection using a dynamic derived from the {replicator equation} \citep{taylorEvolutionaryStableStrategies1978,schusterReplicatorDynamics1983, nowakEvolutionaryDynamicsExploring2006}. Although originally formulated in evolutionary biology \citep{priceSelectionCovariance1970, bomzeLotkaVolterraEquationReplicator1983, pageUnifyingEvolutionaryDynamics2002}, the replicator equation provides a general description of adaptation that extends naturally to cognition and learning \citep{harperReplicatorEquationInference2010, czegelBayesDarwinHow2022, borgersLearningReinforcementReplicator1997, freundDecisionTheoreticGeneralizationOnLine1997, hennesNeuralReplicatorDynamics2020}, making it a natural framework for modeling language evolution.

Finally, as part of our evolutionary modeling, we will follow \citet{frankeVaguenessImpreciseImitation2018} in assuming that there are two unbounded and well-mixed populations---a Sender population and a Receiver population---and that types within populations of Senders and Receivers are represented as particular state-signal pairs. In this model, while individual agents follow  deterministic strategies, their collective behavior can be described in probabilistic terms. Thus, $S(w | x_o)$ represents the probability a Sender randomly sampled from the population uses signal $w$ upon observing $x_o$, and $R(\hat{x}_o | w)$ represents the probability that a randomly sampled Receiver interprets signal $w$ as $\hat{x}_o$. These probabilistic mappings can be used to jointly define a single language with mixed strategy profiles. Taking this particular formalization of strategy evolution, moreover, allows us to study how individuals change their behavior at specific choice points (i.e., upon observing a state or receiving a signal). This is in contrast to prominent previous work modeling language evolution using replicator dynamics, which has assumed changes at the level of entire vocabularies \citep{nowakEvolutionLanguage1999, pawlowitschWhyEvolutionDoes2008} or grammars \citep{komarovaReplicatorMutatorEquation2004, nowakEvolutionaryDynamicsExploring2006}. This has the welcome benefit, as \cite{frankeVaguenessImpreciseImitation2018} note, this approach is more compatible with ideas of imitation and social learning than with biological evolution among agents inflexibly executing their innate behavior. In the next section, we explain in more detail how this dynamic can be used to build an integrative model of how language might evolve culturally from communicative pressures.

\section{Integrating efficient communication and signaling game dynamics}
\label{sec:unified-framework}

Our overarching objective is to provide a concrete and mechanistic model of how human languages might evolve under pressure to efficiently compress meanings into words, as predicted by IB. To do this, we bridge two major theoretical frameworks: the IB framework and the evolutionary dynamics of signaling games. This also allows us to clearly evaluate whether notions of optimality in EGT (success in signaling games) may relate to information-theoretic optimality.  Below, we describe the result of this unification and how the population dynamics of noisy social imitation can be used to test an evolutionary model of efficient semantic systems against IB's theoretical predictions.

\subsection{A unified communication model}
\label{sec:model}

\begin{figure}
    \centering
    \includegraphics[
    width=0.75\linewidth
    ]{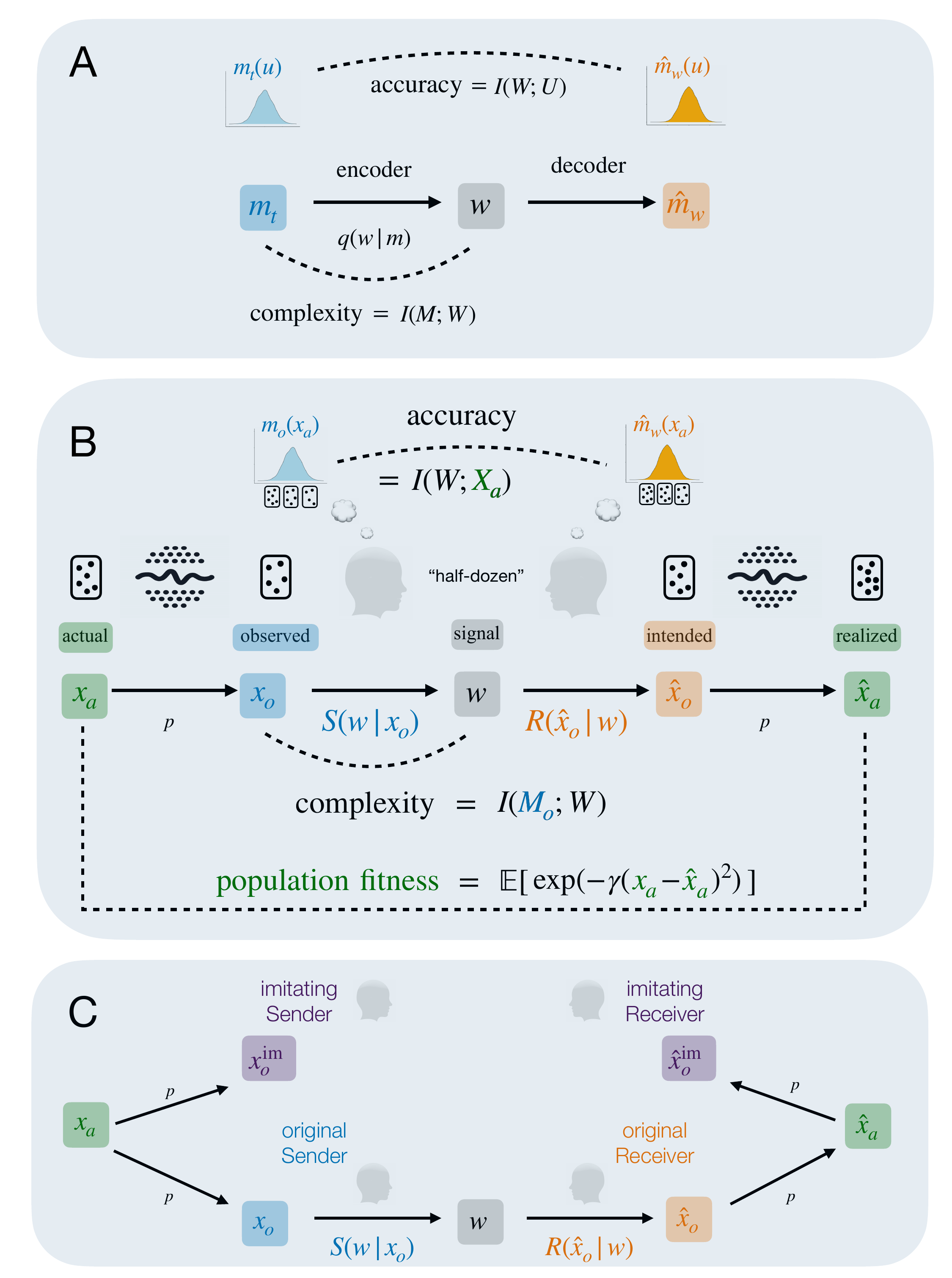}
    \caption{
    \textbf{A.} The IB communication model of \cite{zaslavskyEfficientCompressionColor2018}, see \Cref{sec:ib} for details.
    \textbf{B.} Communication model unifying noisy sim-max games with the IB framework. We illustrate our model on a semantic domain similar to numerals, and depict states as cards with varying quantities of dots, and meanings as (conditional) belief distributions about these states. We denote the (symmetric) noisy perception and interpretation of states as $p$. 
    \textbf{C.} The imprecise conditional imitation dynamic \citep{frankeVaguenessImpreciseImitation2018}. See main text and Appendix \ref{app:dynamics} for details.\\
    }
    \label{fig:main_diagram_figure}
\end{figure}

\Cref{fig:main_diagram_figure}B presents a unified communication model that integrates the noisy sim-max game of of \cite{frankeVaguenessImpreciseImitation2018} with the information-theoretic framework for semantic systems of \cite{zaslavskyEfficientCompressionColor2018}. As before, nature chooses a state of the world $x_a$ with some probability, $p(x_a)$. A Sender $S$, who is randomly sampled from the population, observes $x_o\sim p(x_o | x_a)$, and aims to communicate this observation to a Receiver.  Due to confusability of similar states, Sender forms a belief representation $m_o(x_a)\equiv p(x_a|x_o)$ about the actual state of the world and produces a word $w\sim S(w|x_o)$ to communicate this belief.  That is, the belief state $m_o$ corresponds to the communicative target and the communicative need distribution in this unified model is therefore given by $p(m_o) = \sum_{x_a} p(x_a)p(x_o|x_a)$. A Receiver $R$, who is also randomly sampled from the population, then observes $w$ and based on that aims to infer  the Sender's observation $\hat{x}_o\sim R(x_o|w)$ and belief representation $\hat{m}_w(x_a)$. The Receiver actually realizes state $\hat{x}_a$ with probability $p(\hat{x}_a | \hat{x}_o)$ as the final reconstruction of $x_a$. The payoff for Sender and Receiver is given by the similarity of states $\mathrm{sim}(x_a, \hat{x}_a)$. The population fitness of the language is the expected payoff, i.e. the expected similarity of actual and realized states, given the Sender and Receiver populations.

\subsection{Communicative efficiency}

How does the notion of efficiency apply in general to this communication model? We begin by considering the mixed strategy profile of the Sender population $S(w|m_o)$, which is conditioned on a belief state $m_o(x_a)$, and treating it as a stochastic encoder. This allows us to quantify the complexity $I_S(M_o; W)$ of the population's semantic system. We then use the same mixed Sender strategy profile to characterize the accuracy $I_S(W;X_a)$ that is achievable in principle, using a profile that corresponds to a Bayesian response to the Sender population, $\hat{m}_w(x_a) = \sum_{m_o} S(w | m_o) m_o(x_a)$.  While expected utility depends on the actual population of Receivers, we consider the Bayesian receiver in order to facilitate direct comparison to the optimal systems predicted by IB, which assumes a Bayesian decoder. A semantic system that emerges from Senders playing the noisy sim-max game will be communicatively efficient to the extent it minimizes $\mathcal{F}_{\beta}[S] = I_S(M_o; W) - \beta I_S(W;X_a)$ for some $\beta$. This quantity will always be bounded from below by the value of $\mathcal{F}_{\beta}$ from the true optima, i.e. by the solutions to \Cref{eq:langrangian}. Note that, as in IB, our overall communication model applies generally to any well-specified semantic domain: the noisy signaling game and the IB framework are both formulated without reference to any specific assumptions about the psychological similarity of states, or prior distribution (i.e., communicative need) of states.

\subsection{Evolutionary dynamics via imprecise imitation}

The second contribution of our framework is to investigate the evolution of efficiency with respect to a concrete, dynamic model of cultural evolution. To do this, we explore how semantic systems may emerge from a population of noisy sim-max game players based on general principles of frequency-dependent selection using the \emph{imprecise conditional imitation dynamic}. This dynamic was introduced by \citet{frankeVaguenessImpreciseImitation2018} to extend the replicator equation to signaling populations whose agents perceive states with uncertainty. A schematic description of the dynamic is shown in \Cref{fig:main_diagram_figure}C. 

The mechanism of imprecise social imitation is modeled as follows. Given that each agent can be described as possessing a repertoire of pure strategies---i.e., deterministic pairings of meanings and words---then at each time-step in evolution, an agent may revise their pure strategy by observing and imitating other agents' behavior. Suppose the actual state of the world is $x_a$. An imitating Sender will noisily perceive this as $x_o^{im}$, and has the option of what signal they will send in this perceived state. What signal is sent by the imitator will depend on the probability of having observing $w$ sent by another Sender, who may have noisily perceived state $x_a$ as $x_o$. Likewise, if $w$ was the signal received by an imitating Receiver, then the interpretation that this Receiver assigns to this signal depends on the state $\hat{x}_a$ finally realized by the Receiver agent being imitated, and the imitator's noisily perceived version $\hat{x}_o^{im}$. The formal details of this dynamic and a comprehensive summary of our notation is contained in Appendix \ref{app:dynamics}. For brevity, we simply summarize the discrete-time updates below, informally.

The frequency of word $w$ being used to communicate state $x^{\mathrm{im}}_o$ at the next time step grows in proportion to (i) the expected frequency of $w$ given that $x^{\mathrm{im}}_o$ was observed by some (randomly sampled) imitating Sender, times (ii) the expected utility of this word relative to the population of Receivers. Likewise, the frequency of interpreting $w$ as meaning $\hat{x}^{\mathrm{im}}_o$ grows in proportion to (i) the probability that $\hat{x}^{\mathrm{im}}_o$ is observed by a randomly sampled agent imitating a randomly sampled Receiver that actually realized $\hat{x}_a$ after receiving word $w$, times (ii) the expected utility of the interpretation $\hat{x}^{\mathrm{im}}_o$ given $w$. Though tracing the full graphical model of the game dynamics of noisy imitation here can quickly become complex, the mechanism underlying our model is grounded in two simple ideas: (1) Agents are minimalistic, rather than cognitively sophisticated, and (2) communicative behaviors are imitated according to frequency-dependent fitness. Together, these two principles allow us to study a concrete, interpretable potential mechanism for the evolution of IB-efficiency in language.

\subsection{Predictions}

If the dynamics of imprecise social imitation in signaling games constitute a useful model of the evolution of natural-like semantic systems, then we should expect the systems emerging from the dynamical model to align with the predicted IB optimal systems. Specifically, we should expect the emergent systems to (1) lie close to the IB bound, and have small loss in efficiency compared to optimal systems, and (2) show category structure similar to the optimal IB systems. In the remainder of this paper, we show how to test these predictions, using a concrete semantic domain and an array of simulations.

\section{Test case: Numerosity}
\label{sec:test_case}

To begin to explore the properties of our model, we consider a synthetic domain corresponding to a linear, one-dimensional semantic space. Specifically, the world states for Sender and Receiver are $n$ numbers $\mathcal{X} = \{0, 1, \dots, n-1\} $, and $n$ available words. In all our simulations, we set $n=100$. Conceptually, this corresponds to a setting in which communicative agents must convey scalar magnitudes under noisy perceptual or cognitive constraints. In principle, this state space could represent a variety of conceptual or perceptual continua, such as brightness, temperature, spatial position, or an idealized mental number line. Of particular relevance to prior work on efficiency in language is the interpretation of this space as one of approximate, idealized numerosity. 

Numerosity presents a good test-bed for our evolutionary model, as such systems have been studied in the context of semantic typology \citep{xuNumeralSystemsLanguages2014, denicRecursiveNumeralSystems2024}, and also in emergent communication \citep{carlssonLearningApproximateExact2021}, but not yet from the perspective of IB.\footnote{A comprehensive empirical evaluation of our framework on actual natural language numeral systems would also involve a power-law distributed communicative need distribution, additional constraints on representation (e.g., empirically estimated similarities of numerosity, and mechanisms for recursion), and evaluating on attested typological data. Our goal in this paper, however, is not to provide a fine-grained diachronic model of one semantic domain, but rather to outline a more generally applicable, proof-of-concept dynamical model to explain the evolution of near IB-optimal category systems.} The resulting IB systems will bear similarity to natural language numeral systems; at low complexity ($\beta \rightarrow 1$), IB solutions resemble approximate numeral systems, while as $\beta \rightarrow \infty$, the IB solution is a bijective mapping from meanings to words and models exact numerosity. This is possible in principle because there are equally many words as meanings available in the communication game; there is no guarantee, however, that the imitation dynamic will carry the population to any particular IB solution, including the bijective mapping that corresponds to exact numerosity.

In our evolutionary model, successful communication depends jointly on pragmatic conventions (how precisely meanings are expected to be conveyed) and on perceptual constraints governing how similar or confusable nearby magnitudes are. The following subsections specify how these factors are instantiated in our synthetic domain through the payoff structure and form of state confusability.

\subsection{Payoff structure and confusion of states}
\label{sec:payoffs}

\paragraph{Utility via pragmatic standards}
For our communicative utility function, we follow \citep{frankeVaguenessImpreciseImitation2018, regierWordMeaningsLanguages2015} in using a model from mathematical psychology \citep{shepardStimulusResponseGeneralization1957, nosofskyAttentionSimilarityIdentification1986} to model the similarity of meanings. Formally, the similarity between state $x, x' \in \mathcal{X}$  is given by:
\begin{equation}
  \label{eq:similarity}
  \text{similarity}_{\gamma}(x, x') =  \exp (-\gamma ~ (x - x')^2 ) ~~ \text{s.t}. ~~ \gamma \geq 0.
\end{equation}
Then we define utility $u_{\gamma}(x, x')$ simply to be the similarity between $x$ and $x'$:
\begin{equation}
    \label{eq:utility}
    u_{\gamma}(x,x') = \text{similarity}_{\gamma}(x, x').
\end{equation}
In \Cref{eq:similarity},  $\gamma$ is an imprecision parameter representing a tolerable level of pragmatic slack. The minimum standard of precision (i.e., maximum pragmatic slack) is assumed when $\gamma = 0$, which lets all pairs of Sender and Receiver meanings receive identical payoff. Perfect discrimination between states is enforced when $\gamma \rightarrow \infty$, in which only perfect guesses by Receiver yield nonzero payoff. We then define the population fitness for Sender and Receiver as the expected similarity between actual states of Nature and finally reconstructed states by Receiver:

\begin{align}
    \label{eq:team_eu}
    \mathbb{E}_{\mathrm{Pr},S,R}[u_{\gamma}(x_{a}, \hat{x}_a)] &= \left(\sum_{x_{a}} \sum_{x_{o}} \mathrm{Pr}(x_a) p(x_{o} | x_{a}) \sum_{w} S(w | x_{o}) \sum_{\hat{x}_{o}} R(\hat{x}_{o} | w) \sum_{\hat{x}_{a}} p(\hat{x}_{a} | x_{o}) \right)\cdot u_{\gamma}(x_{a}, \hat{x}_a) ,
\end{align}
where $p(x_o | x_a) \equiv p(x_a | x_o)$ represents the probability of confusing similar states, defined below, and $\mathrm{Pr}(x_a)$ is a prior over meanings reflecting the need to which each needs to be communicated.

\paragraph{Prior over states}
We assume a uniform prior over states, giving each magnitude equal communicative need. This neutral setting allows us to isolate the role of perceptual noise and pragmatic incentives in shaping efficient systems. However, this assumption can easily be relaxed, and future extensions could introduce non-uniform priors to model realistic distributions of need or frequency.\footnote{In preliminary simulations, replacing this with a power-law distributed prior did not qualitatively change our results. Therefore, to keep our model simple and transparent, we focus on the uniform setting.} To evaluate resulting systems with respect to IB, we considered a uniform prior over meanings identical to the prior over states, i.e. $p(x_a) = p(x_o)$, in order to avoid artifacts due to boundary conditions when deriving $p(x_o)$ from confusion probabilities.

\paragraph{State confusion probabilities} Following \cite{frankeVaguenessImpreciseImitation2018}, we assume that the probability of confusing states depends on their similarity. To do this, we renormalize the similarity function (\Cref{eq:similarity}) to probabilities and fix a perceptual certainty parameter $\alpha$, representing the level of noise in the game dynamics. We define the probability of confusing state $x$ for state $x'$ as proportional to their similarity:
\begin{equation}
    \label{eq:confusion}
     p_{\alpha}(x' \mid x) = \frac{ \text{similarity}_{\alpha}(x , x') }{ \sum_{x'} \text{similarity}_{\alpha}(x , x') }.
\end{equation}

\noindent Note that utility and state confusion are defined using the same similarity function, but their parameters---$\gamma$ and $\alpha$, respectively---are set independently. In all of our analyses, we fix $\alpha = 0.5$ (i.e., so that the state confusion function is Gaussian with width 1 centered at the actual state).

\section{Results}
\label{sec:results}

\begin{figure}[t!]
    \centering
    \includegraphics[width=0.99\linewidth]{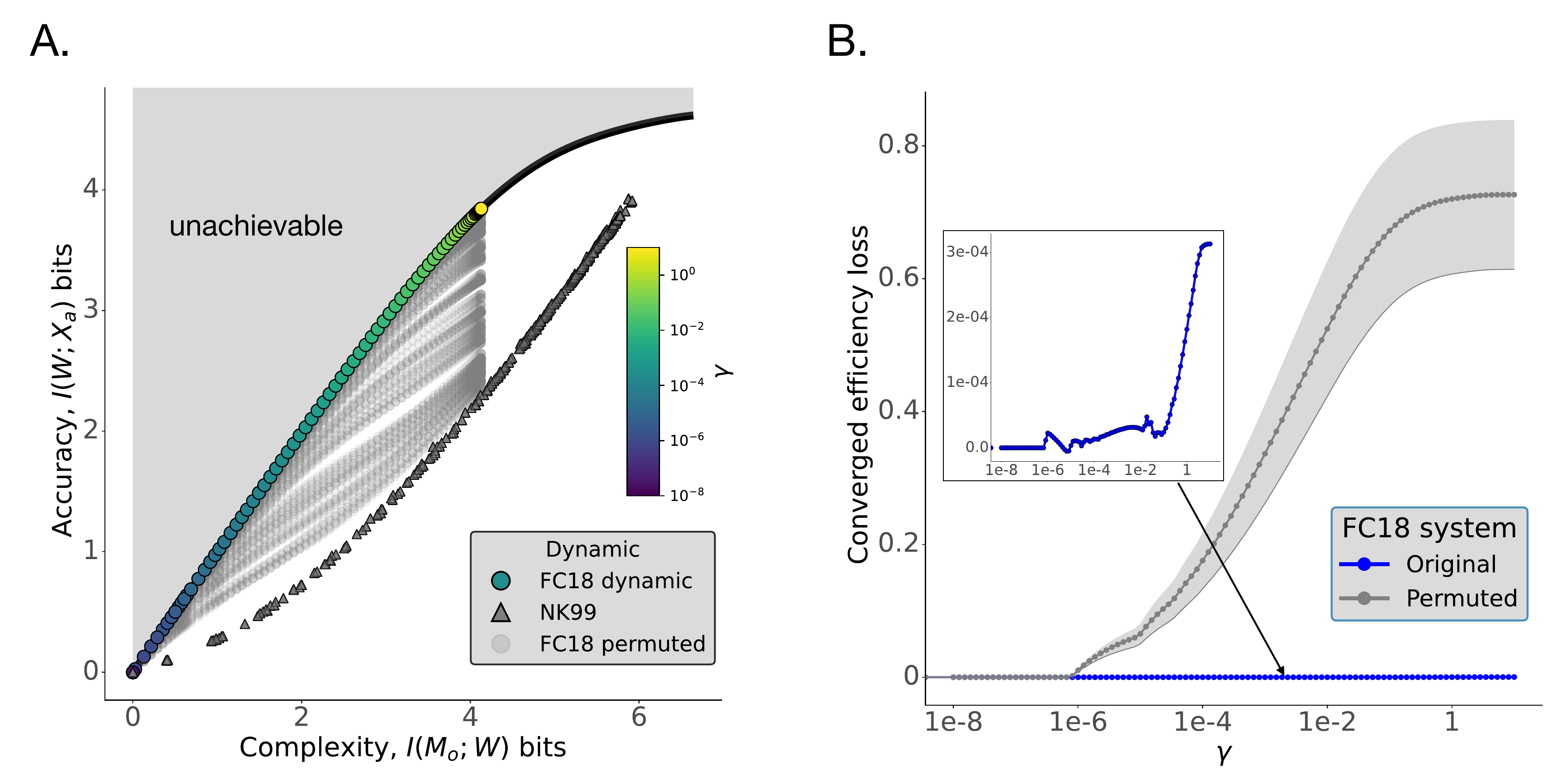}
    \caption{ Emergent efficient semantic systems.
    \textbf{A}: Converged systems on the information plane, colored by the pragmatic standard of precision $\gamma$ in the sim-max game. Circles denote systems evolved from our model, derived from \cite{frankeVaguenessImpreciseImitation2018}, abbreviated as FC18. See \href{https://www.dropbox.com/scl/fi/vatlo672rmdqzf49s8uf8/trajectory_on_bound.mp4?rlkey=ilnbzte18kigwdcifgt4hlrsy&st=wco4ozim&dl=0}{Movie~1} visualizing the dynamics on the plane. Gray circles mark the result of permuting the systems evolved from the FC18 dynamics (see Appendix \ref{app:baselines}). Gray triangles correspond to emergent systems evolved from a dynamic proposed in \cite{nowakEvolutionLanguage1999}, abbreviated as NK99, which we consider as another baseline. 
    \textbf{B}: Game pragmatic standard of precision ($\gamma$, \Cref{eq:similarity}) vs. efficiency loss ($\epsilon$, \Cref{eq:epsilon}) of converged systems. Mean scores for systems emerging from our model are depicted in blue while mean scores for permuted counterparts are in gray. Shaded regions correspond to the $95\%$ confidence interval. The inset shows a zoomed-in view of the emergent systems' scores only.\\
    }
    \label{fig:compare_on_bound}
\end{figure}
\begin{figure}[t!]
    \centering
    \includegraphics[width=0.99\linewidth]{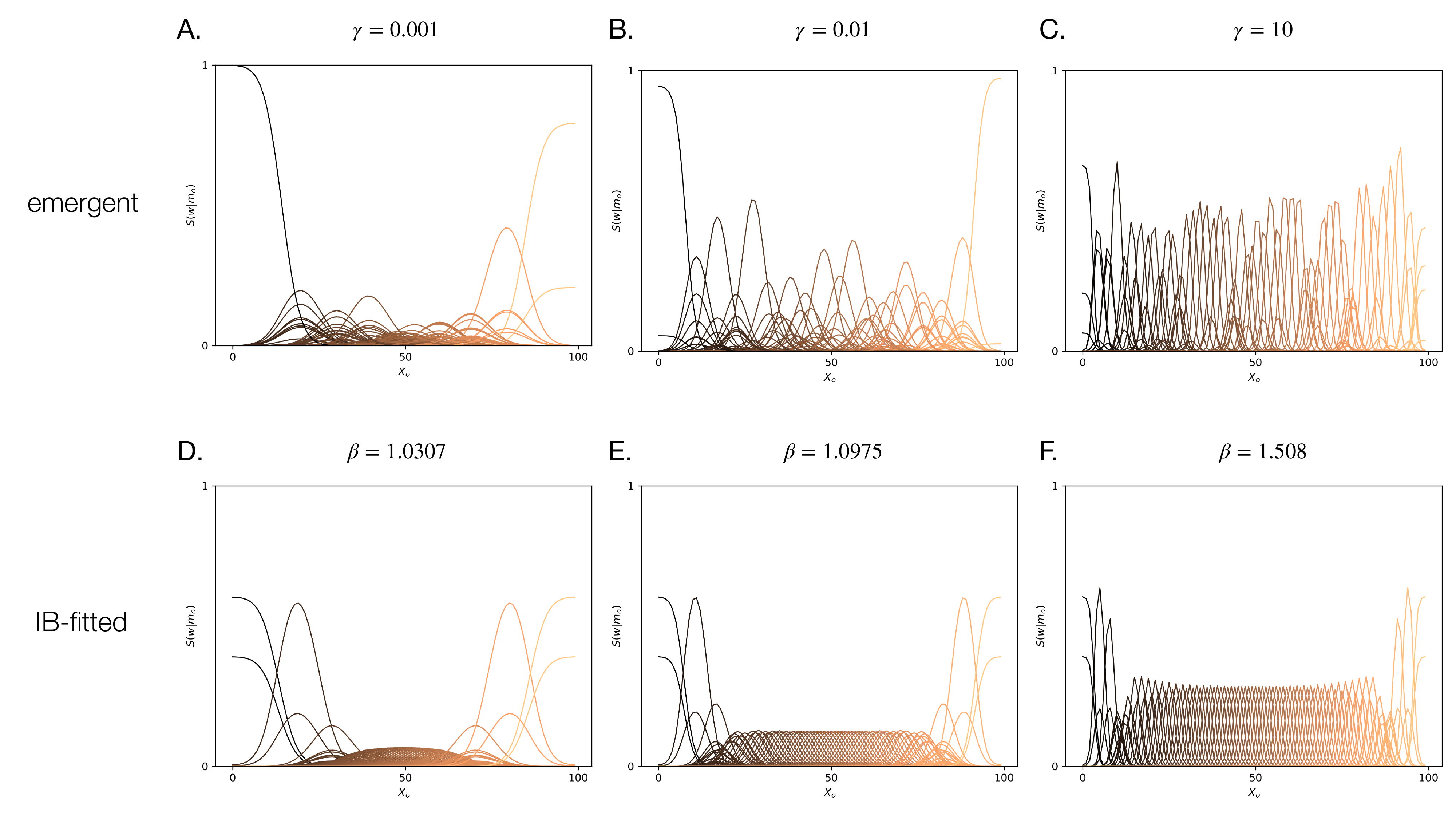}
    \caption{
        Sample of converged emergent systems and their optimal counterparts. \textbf{A}-\textbf{C}: 1-dimensional mode maps of converged system for $\gamma=0.001, 0.01, 10$ (\Cref{eq:similarity}). In each plot, the $x$-axis denotes the meaning space $\mathcal{X}_o = \{0, \dots, 99\}$, the $y$-axis denotes probability $S(w|m_o)$, and lines are the modal word $w$ (out of $100$ possible words) used for each meaning $m_o = x_o$. The color of each line corresponds to the average meaning that the word is used to communicate, with black corresponding to $x_1$ and bright orange corresponding to $x_{100}$. \textbf{D}-\textbf{F}: Mode maps for the optimal counterparts of the emergent systems in \textbf{D}-\textbf{F}, fitted by efficiency loss $\epsilon$ (\Cref{eq:epsilon}). See \href{https://www.dropbox.com/scl/fi/igdumose4bi0qx13vcfzj/system_movie.mov?rlkey=s3tx4mjylp86b8nemlp13zcni&st=x03po5qp&dl=0}{Movie~2} for the evolution of the system in \textbf{A}.\\
    }
    \label{fig:fits}
\end{figure}
\begin{figure}[t!]
    \centering
    \includegraphics[width=0.99\linewidth]{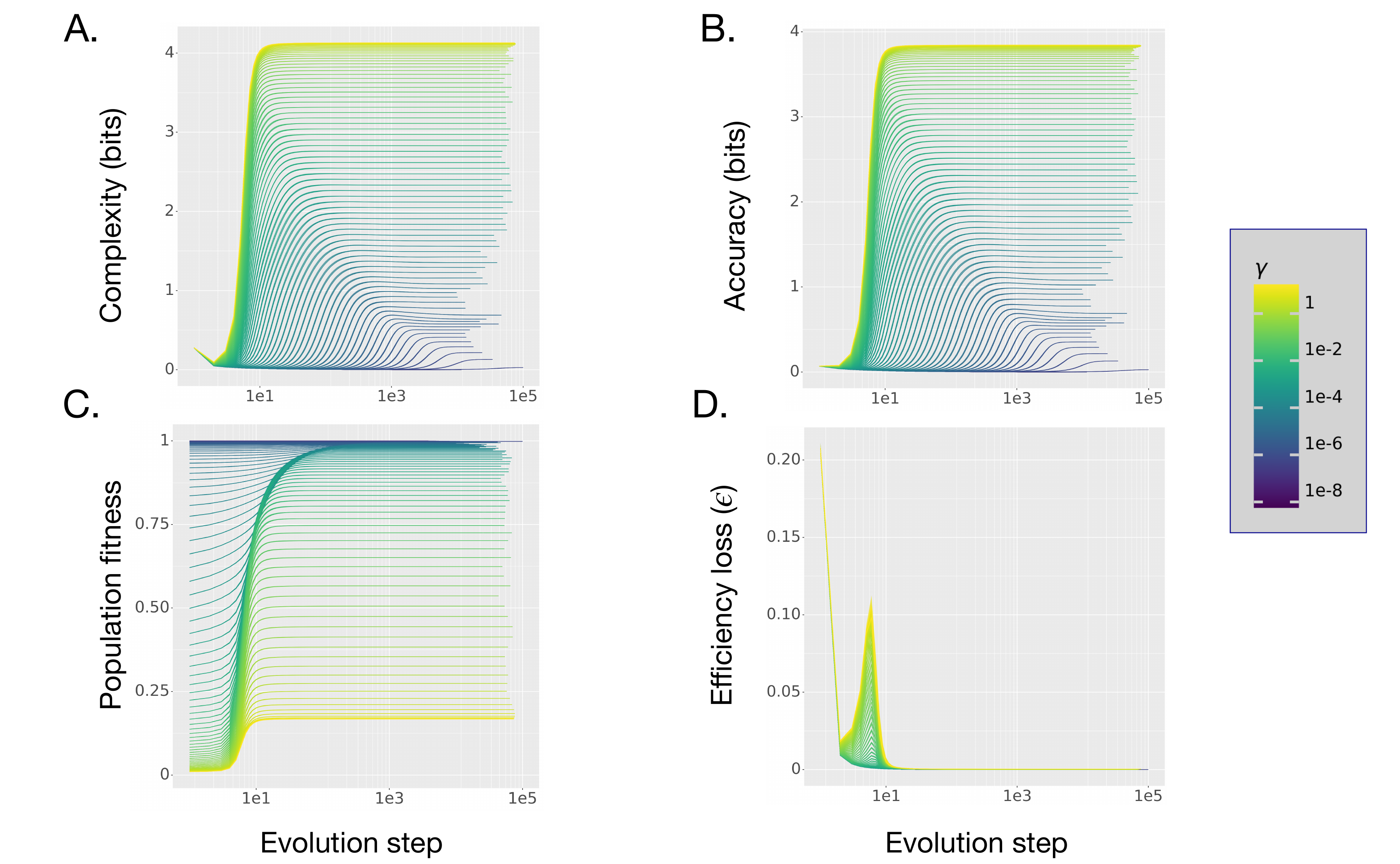}
    \caption{
    Complexity (A), accuracy (B), population fitness (C), and efficiency loss (D), relative to game pragmatic standard of precision ($\gamma$) across discrete time step of simulated evolution. Lines in each plot show mean (with 95\% CIs) across eight random seeds. Color corresponds to $\gamma$ (\Cref{eq:similarity}).\\
    }
    \label{fig:metrics}
\end{figure}

\subsection{Fixed points}

We simulated the imprecise imitation dynamic for up to $10^5$ time steps, measuring team expected utility, complexity, accuracy, and deviation from optimality defined by $\varepsilon = \min_{\beta} \{ \frac1\beta (\mathcal{F}_{\beta}[q] -  \mathcal{F}_{\beta}^*)\}$ where $\mathcal{F}_{\beta}^*$ is the optimal value of $\mathcal{F}_{\beta}$ from \Cref{eq:langrangian}~\citep[see][and Appendix \ref{app:ib} for more details]{zaslavskyEfficientCompressionColor2018}. We find that this dynamic produces communication systems that are remarkably close to the IB bound of efficient compression (\Cref{fig:compare_on_bound}A). Emergent systems achieve near-optimal trade-offs between complexity and accuracy, outperforming both randomly permuted controls and systems evolved under a finite-population replicator–mutator dynamic \citep{nowakEvolutionLanguage1999}. For full methodological details on our baselines, see Appendix \ref{app:baselines}.

To ensure that these findings generalize across a wide range of communicative environments, we systematically varied the pragmatic standards parameter $\gamma$ which determines the sharpness of game payoffs, across $100$ logarithmically spaced values in the range $[10^{-8}, 10]$, which we found to span the empirically achievable saturation points of complexity, accuracy and utility for emergent systems. We also varied initial conditions across eight random seeds, yielding a total of $800$ runs. Almost all systems converged well before our limit of $10^5$ steps.\footnote{
Convergence was defined to be when the sum of absolute elementwise differences between the matrices for successive population distributions dropped below $10^{-5}$, which typically occurred before $40,000$ steps. All simulations converged except for one setting: $\gamma \approx 1e-6$. Near this value, and also near $\gamma \approx 1e-2$, convergence was slower and efficiency loss locally higher (\Cref{fig:compare_on_bound}B), suggesting possible critical behavior near these values.}

Pragmatic standards shape the position of emergent systems on the information plane in the expected manner: games that reward finer distinctions (higher values of $\gamma$) yield more complex and accurate meaning systems. This systematic relationship between pragmatic standards and position on the complexity-accuracy tradeoff is visible in (\Cref{fig:compare_on_bound}A), and we observe quantitatively a smooth gradient for this trend across intermediate values ($\rho_{\text{complexity}} \approx \rho_{\text{accuracy}} \approx 0.99$, $p=0.0$). 

In addition, \Cref{fig:fits} shows that emergent systems show similar categorization structure to IB-optimal systems. Consistent with the pattern of complexity-accuracy trade-offs we observe, we find that across low, medium, and high values of $\gamma$, the best-fitting (by $\epsilon$) IB systems correspond to solutions at progressively higher $\beta$. We observe a high Spearman rank correlation between the two parameters ($\rho = 0.99$, $p=0.0$) when paired based on this fitting (see \Cref{app:gamma_vs_beta}, \Cref{fig:gamma_vs_beta}). These findings suggest that the pragmatic standard of precision in communication strongly constrains the distribution of near-optimal systems along the IB bound, and is therefore a promising agent-level mechanistic parameter for explaining variation in efficiency trade-offs.

As can be observed in \Cref{fig:compare_on_bound}A, however, the emergent systems do not span the full range of trade-offs. Noisy imitation appears to prevent convergence to a fully bijective mapping between meanings and words, even when such mappings are globally optimal (e.g., when $\gamma=1$). The maximum achievable accuracy in this domain is $I(M_o; X_a) \approx 4.61$ bits, yet the highest accuracy of any emergent system remains below $4$ bits, suggesting a fundamental limit on communicative precision under noisy imitation dynamics. Interestingly, the emergent systems' efficiency loss (deviation from optimality) also increases with $\gamma$, (\Cref{fig:compare_on_bound}B; $\rho = 0.93$, $p=0.0$), indicating that as pressure for precise communication grows, IB-optimal solutions become harder for the imitation dynamic to approximate. 

Nevertheless, the emergent systems do achieve near-optimal efficiency: they approximately minimize complexity for their achieved accuracy values, and are more efficient than baseline counterpart systems. Crucially, this emergent efficiency is not an a priori feature of the model, but rather an empirical finding, as there is nothing in the evolutionary dynamics that guarantees IB-efficiency. In particular, although higher accuracy is expected because strategies yielding greater expected utility are preferentially imitated, it is not obvious why the dynamics should drive systems to also minimize complexity. Intuitively, one possibility is that noisy imitation acts as a regularizer that softens category boundaries, and thereby indirectly reduces complexity.

\subsection{Dynamics}

Over iterations, we observe that expected team utility and communicative accuracy increase, and complexity increases roughly the minimum amount required for each current achieved accuracy, resulting in population systems remaining close to the bound throughout evolution (\Cref{fig:metrics}, see also \href{https://www.dropbox.com/scl/fi/vatlo672rmdqzf49s8uf8/trajectory_on_bound.mp4?rlkey=ilnbzte18kigwdcifgt4hlrsy&st=wco4ozim&dl=0}{Movie~1} and \href{https://www.dropbox.com/scl/fi/igdumose4bi0qx13vcfzj/system_movie.mov?rlkey=s3tx4mjylp86b8nemlp13zcni&st=x03po5qp&dl=0}{Movie~2}). 

Despite this general tendency toward compression, efficiency does not always improve monotonically over time. As shown in \Cref{fig:metrics}D, deviation from optimality remains substantially below chance but exhibits transient fluctuations. In particular, a pronounced spike in efficiency loss occurs during the first $100$ iterations, with the effect becoming more pronounced at higher values of $\gamma$. 

In contrast, the baseline finite-population dynamic we considered from \citet{nowakEvolutionLanguage1999} fails to achieve the levels of efficiency reached by systems evolved under our model, or even the efficiency reached by the systems resulting from random permutations of our model's converged systems. While some runs of the baseline dynamic lead to populations develop vocabularies whose words carry relatively high information about the environment, these systems are inefficient: they do not minimize complexity for the accuracy they achieve.

Taken together, the results of our simulations show that a model of imprecise imitation of communicative behavior can drive a population's vocabulary towards near-optimal compression. Furthermore, our model predicts that both the local, contextual standards of precision in these games, and individuals' tendency to confuse similar states, impose systematic constraints on the specific efficient trade-offs achieved by emergent vocabularies. 

\section{Discussion}
\label{sec:discussion}
In this paper, we set out to provide a mechanistic explanation of how efficient communication can arise in a population of language speakers, and to investigate whether previously studied notions of success in communication from game theory may be related to information-theoretic optimality. To do this, we unified a mathematically precise and empirically supported account of efficiency in semantic systems based on the IB principle \citep{zaslavskyEfficientCompressionColor2018}, with a model of the emergence of vague meaning from evolutionary game theory \citep{frankeVaguenessImpreciseImitation2018}. Our simulations demonstrates that a replicator dynamic of noisy social imitation can drive a communicative population's vocabulary towards the theoretical bounds of efficiency. We also found that both the local, contextual standards of precision in these games and individuals' tendencies to confuse similar states impose systematic constraints on the specific tradeoffs achieved by emergent vocabularies. These results suggest that evolutionary game dynamics can provide a mechanistic basis for the evolution of vocabularies with information-theoretically optimal and empirically attested properties. 

Our framework also offers conceptual and theoretical foundations for further studying how game-theoretic and information-theoretic principles relate in language evolution. For example, in our framework, we have build on a game-theoretic approach that models the Sender and Receiver populations as emergent distribution of many individuals' pure strategies, which are not explicitly driven by any pressure for efficiency. One consequence of this is that even if the emergent population-level system achieve efficient compression, this does not necessarily imply that each individual in the population must be explicitly guided by efficiency. Interestingly, however,~\cite{Imel2025Iterated} recently found empirical evidence suggesting that individuals do exhibit an inductive learning bias toward IB-efficient semantic system. This opens the question of why such a bias might emerge at the individual level, while, as we have shown, it is not necessary for population-level efficiency. 
Another important direction for future research is to further explore the mathematical relationships between evolutionary dynamics and information theoretic optimization. 

\section*{Acknowledgments}

We thank Shane Steinert-Threlkeld for valuable feedback on earlier versions of this work. We are also grateful to the members of the Departments of Logic and Philosophy of Science and Language Science at UC Irvine, members of the MIT Computational Psycholinguistics Lab, as well as participants at the NeurIPS 2023 InfoCog Workshop and the CogSci 2023 conference for helpful discussions during the development of this project.


\bibliography{references}

\appendix

\section{The replicator equation and the imprecise conditional imitation dynamic}
\label{app:dynamics}

The general, multiplicative form of the discrete time replicator equation is:
\begin{align}
\label{eq:discrete_time_replicator_equation}
    p^{(t+1)}_{i} &= p^{(t)}_{i} \left[ \frac{f_{i}(p^{(t)})}{\phi(p^{(t)})}  \right],
\end{align}
where $p \in \mathbb{R}^n$ is a vector of the distribution of $n$ possible types in the population, $p_i$ is the proportion of type $i \in [n]$ in the population, $f_i(p)$ is the expected fitness type $i$ relative to the population, and $\phi (p)=\sum _{j=1}^{n}{p_{j}f_{j}(p)}$ represents the mean fitness of the population. Intuitively, \Cref{eq:discrete_time_replicator_equation} says that if at time $t$ a type has greater fitness than average, its proportion in the population will increase in the next time step.

\begin{figure}[h!]
    \centering
    \includegraphics[width=.8\textwidth]{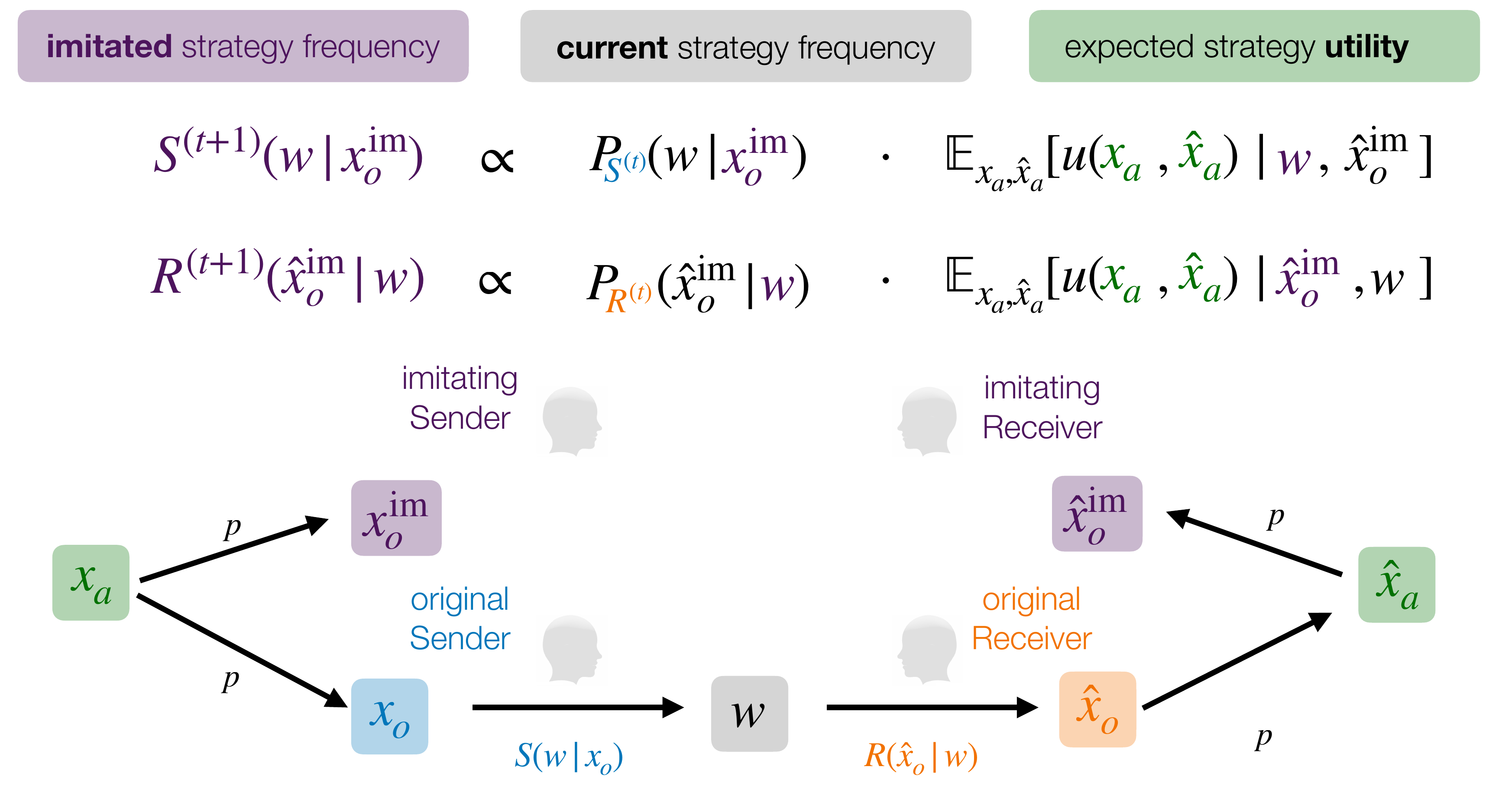}
    \caption{
    The imprecise conditional imitation dynamic \citep{frankeVaguenessImpreciseImitation2018}. The frequency of word $w$ being used to communicate state $x^{\mathrm{im}}_o$ at the next time step grows in proportion to (i) the expected frequency of $w$ given that $x^{\mathrm{im}}_o$ was observed by some (randomly sampled) imitating Sender, times (ii) the expected utility of this word relative to the population of Receivers. Likewise, the frequency of interpreting $w$ as meaning $\hat{x}^{\mathrm{im}}_o$ grows in proportion to (i) the probability that $\hat{x}^{\mathrm{im}}_o$ is observed by a randomly sampled agent imitating a randomly sampled Receiver that actually realized $\hat{x}_a$ after receiving word $w$, times (ii) the expected utility of the interpretation $\hat{x}^{\mathrm{im}}_o$ given $w$. See \Cref{eq:eu_sender,eq:imitation_prob_sender,eq:eu_receiver,eq:imitation_prob_receiver,eq:sender_update,eq:receiver_update} for the details of these conditional expectations. \\
    }
    \label{fig:dynamic}
\end{figure}

\citeauthor{frankeVaguenessImpreciseImitation2018}'s imprecise conditional imitation dynamic, visualized as a graphical model in \Cref{fig:dynamic}, describes how social imitation and communication interact in a stochastic, adaptive manner.\footnote{To be precise, this dynamic describes events probabilistically, but like the standard replicator equation it is a deterministic process at the level of modeling.} The interpretation of this process is as follows. At each time-step in evolution, an agent may revise their pure strategy, based on observing and imitating other agents' behavior. Suppose the actual state of the world is $x_a$. An imitating Sender will noisily perceive this as $x_o^{im}$, and has the option of what signal they will send in this perceived state. This will depend on the probability of having observing $w$ sent by another Sender, who may have noisily perceived state $x_a$ as $x_o$. Likewise, if $w$ was the signal received by an imitating Receiver, then the interpretation that this Receiver assigns to this signal depends the state $\hat{x}_a$ finally realized by a the Receiver agent being imitated, and the imitator's noisily perceived version $\hat{x}_o^{im}$. \Cref{tab:notation} summarizes our notation.

\begin{table}[t!]
    \centering
    \begin{tabular}{c l}
        \toprule
         Notation  & Interpretation \\
         \midrule
         $x_a$  & actual state \\
         $x_o$  & Sender's observed state \\
         $x^{im}_o$ & an imitating Sender's observed state \\
         $m_o(x_a)\equiv p(x_a|x_o)$  & observation noise / belief state \\
         $p(m_o)$ & communicative need distribution \\
         $w$ & signal \\
         $S(w|m_o) \equiv S(w|x_o)$ & probability that a random Sender uses strategy $w$ in $x_o$ \\
         $\hat{x}_o$ & Receiver's intended reconstruction state \\
         $\hat{x}^{im}_o$ & an imitating Receiver's observed state \\                           
         $R(\hat{x}_o|w)$ & probability that a random Receiver uses strategy $\hat{x}_o$ in $w$ \\
         $\hat{x}_a$ &  Receiver's realized reconstruction state \\
         $\hat{m}_w(x_a) = \sum_{x_o} R(x_o|w) p(x_a|x_o)$  & Receiver's reconstruction of Sender's belief state  \\
        \bottomrule
         \end{tabular}
    \caption{Mapping between our notation, \citet{frankeVaguenessImpreciseImitation2018}'s notation, and \citet{zaslavskyEfficientCompressionColor2018}'s notation.}
    \label{tab:notation}
\end{table}

As an instance of the replicator dynamic, our model assumes that the frequency of a signaling behavior evolves according to its current frequency and expected utility relative to other behaviors' frequencies in the current population. Thus evolution is driven by not only the communicative utility of signaling strategies, but also how these strategies interact with the current landscape of existing strategies. Since we are modeling evolution in sim-max games, the utility of Sender and Receiver's strategies are defined in terms of how well they jointly maximize similarity between states $x_a, \hat{x}_a \in \mathcal{X}$.  Below, we describe in detail the discrete-time update equations of the imprecise conditional imitation dynamic.

\subsection{Modeling agent utility and imitation probabilities}

\paragraph{Sender's expected utility}

The expected utility for a Sender's strategy is defined in terms of the utility of sending $w$ in state $x_a$ given that $x_o$ was observed, with respect to the current Receiver population $R$:
\begin{align}
    \label{eq:eu_sender}
    \mathbb{E}_{x_a, \hat{x}_a}\left[ u(x_a, \hat{x_a}) | w, x_o \right] &= \sum_{x_a,  \hat{x}_a, \hat{x}_o} p(x_a|x_o) R(\hat{x}_o | w) p(\hat{x}_a | \hat{x}_o)  \cdot u(x_a, \hat{x}_a).
\end{align}
To interpret this expectation, consider the following process. Nature first generates a state of the world, $x_a \sim p(x_a)$, which the Sender noisily observes as $x_o$. The first probability in the summation is the probability that $x_a \sim p(x_a|x_o)$ was the true state. Next, $\hat{x}_o \sim R(\hat{x}_o | w)$ is the Receiver's interpretation after observing $w$. Finally, $\hat{x}_a \sim p(\hat{x}_a | \hat{x}_o)$ is the Receiver's final reconstruction of the actual state given their interpretation $\hat{x}_o$, capturing decision-making noise on the Receiver's end.

\paragraph{Sender's imitation probability}
The probability that an imitator Sender will use signal $w$ while having observed state $x_{o}^{im}$ is
\begin{equation}
    \label{eq:imitation_prob_sender}
    P_S(w|x_o^{im}) = \sum_{x_a, x_o}  p(x_a|x_{o}^{im}) p(x_o|x_a) S(w|x_o),
\end{equation}
where $x_a \sim p(x_a | x_o^{im})$ is drawn from the distribution over actual states given that $x_o^{im}$ was observed by the imitating Sender, and $x_o \sim p(x_o | x_a)$ is drawn according to the probability that the imitated Sender will confuse $x_a$ for $x_o$.

\paragraph{Receiver's expected utility}
Parallel to the Sender, The Receiver's fitness function is identified with the utility of guessing $\hat{x}_o$ after receiving $w$, with respect to the current Sender population $S$:
\begin{align}
    \label{eq:eu_receiver}
  \mathbb{E}_{x_a, \hat{x}_a} \left[ u(x_a, \hat{x}_a) | w, \hat{x}_o \right] &= \sum_{x_a, \hat{x}_a} S(x_a | w) p(\hat{x}_a | \hat{x}_o) \cdot u(x_a, \hat{x}_a),
\end{align}
where $x_a \sim S(x_a | w)$ is the state $x_a$ chosen by Nature, given the current Sender population, and $\hat{x}_a \sim p(\hat{x}_a | \hat{x}_o)$ is the Receiver's final reconstruction given the intended interpretation $\hat{x}_o$. 

\paragraph{Receiver imitation probability}
The probability that an imitator will observe and adopt interpretation $\hat{x}_o^{im}$ for signal $w$ is
\begin{align}
    \label{eq:imitation_prob_receiver}
    P_R(\hat{x}_o^{im} | w) = 
    \sum_{\hat{x}_a, \hat{x}_o} p(\hat{x}_o^{im} | \hat{x}_a) p(\hat{x}_a | \hat{x}_o) R(\hat{x}_o|w),
\end{align}
where $\hat{x}_o \sim R(\hat{x}_o | w)$ is the randomly sampled imitated Receiver's intended interpretation of $w$, and $\hat{x}_a \sim p(\hat{x}_a | \hat{x}_o)$ is the imitated Receiver's final reconstruction, given that $\hat{x}_o$ was intended.

Having defined the expected utility for both the Sender and Receiver populations, we can now describe the discrete time update equations for the imprecise conditional imitation dynamic. 

\subsection{Discrete time update equations}

For the update of the Sender population, assume that at time step $t+1$ an imitating Sender observes $x_o^{im}$. Then the probability that the imitator adopts strategy $w$ is:
\begin{equation}
    \label{eq:sender_update}
    \underbrace{
        S^{(t+1)}(w|x_o^{im})
    }_{\text{updated strategy frequency}} 
    \propto \underbrace{ 
        P_S^{(t)}(w|x_o^{im})
    }_{\text{imitated strategy probability}} 
    \cdot ~ \underbrace{ 
        \mathbb{E}_{x_a, \hat{x}_a}\left[ u(x_a, \hat{x_a}) | w, x^{im}_o \right].
    }_{\text{imitated strategy utility}}
\end{equation}
For the Receiver population's update, assume that at time step $t+1$ the imitating Receiver observes $w$. Then the probability that the imitator adopts strategy $\hat{x}_o^{im}$ is:
\begin{equation}
    \label{eq:receiver_update}
    \underbrace{
        R^{(t+1)}( \hat{x}_o^{im} | w)
    }_{\text{updated strategy frequency}} \propto 
    \underbrace{ 
        P_R^{(t)}(\hat{x}_o^{im} | w)
    }_{\text{imitated strategy probability}} 
    \cdot ~ \underbrace{ 
        \mathbb{E}_{x_a, \hat{x}_a} \left[ u(x_a, \hat{x}_a) | w, \hat{x}^{im}_o \right]
    }_{\text{imitated strategy utility}}.
\end{equation}
The proportionality symbols indicate that the right-hand sides must be normalized so that the left hand side is a valid conditional probability distribution. 

To summarize this dynamic and highlight its connection to the bare general form of the discrete time replicator equation in \Cref{eq:discrete_time_replicator_equation}: the characteristic assumption in the replicator dynamic is that a behavior's frequency grows in a population as a function of its current frequency and its expected utility, relative to the frequency of other behaviors in the current population. This is captured by \Cref{eq:sender_update} for the Sender population: we are updating the probability of a sampling a signaling strategy as the probability that an imitating Sender will adopt the strategy, weighted by the utility this strategy is expected to achieve relative to the current Receiver population. Likewise, for the Receiver population, in  \Cref{eq:receiver_update} we are updating the probability of an interpretation strategy by the probability an imitating Receiver will adopt it, weighted by the utility it is expected to achieve relative to the current Sender population. Importantly, this all happens with the possibility that agents will confuse perceptually similar states, which factors into both the probability of an imitator's strategy (\Cref{eq:imitation_prob_sender,eq:imitation_prob_receiver}) and the expected utility of each strategy ( \Cref{eq:eu_sender,eq:eu_receiver}). For a precise derivation of each update rule from the standard replicator equation, we refer readers to Appendix A of \citet{frankeVaguenessImpreciseImitation2018}.

\section{IB theoretical solutions}
\label{app:ib}

\paragraph{Measuring efficiency loss}
As explained in Section \Cref{sec:ib}, we call a language efficient to the extent it minimizes the IB objective function. To make this notion precise, we adopt \cite{zaslavskyEfficientCompressionColor2018}'s metric of {efficiency loss}, or deviation from optimality, to quantify the extent to which any encoder $S$ is suboptimal. This metric $\epsilon_S$ is defined
\begin{equation}
    \label{eq:epsilon}
    \epsilon_S = \frac{1}{\beta_S} \Delta \mathcal{F}_{\beta_S}
\end{equation}
where $\beta_S$ is selected as the $\beta$ that minimizes $\Delta \mathcal{F}_{\beta} = \mathcal{F}_{\beta}[S(w | m_o)] - \mathcal{F}^{*}_{\beta}$. 

\paragraph{Theoretical bound}
To compute the IB bound for our domain, we used the IB method \citet{tishbyInformationBottlneckMethod1999}, in combination with reverse deterministic annealing~\citep{Rose1998}, as described in \cite{zaslavskyEfficientCompressionColor2018} \cite[see also ][Chapter 7]{zaslavskyInformationTheoreticPrinciplesEvolution2020}.
We take $\mathrm{Pr}(x_o)$ to be uniform and $p(x_a | x_o)$ to be equal to the perceptual uncertainty / state-confusion in \Cref{eq:confusion}. To compute the theoretical bound for our synthetic semantic domain, we applied the IB method for $795$ values of $\beta \in [1, 10^{7}]$.

\section{Pragmatic standards vs. IB trade-offs}
\label{app:gamma_vs_beta}

\begin{figure}[h!]
    \centering
    \includegraphics[width=0.5\linewidth]{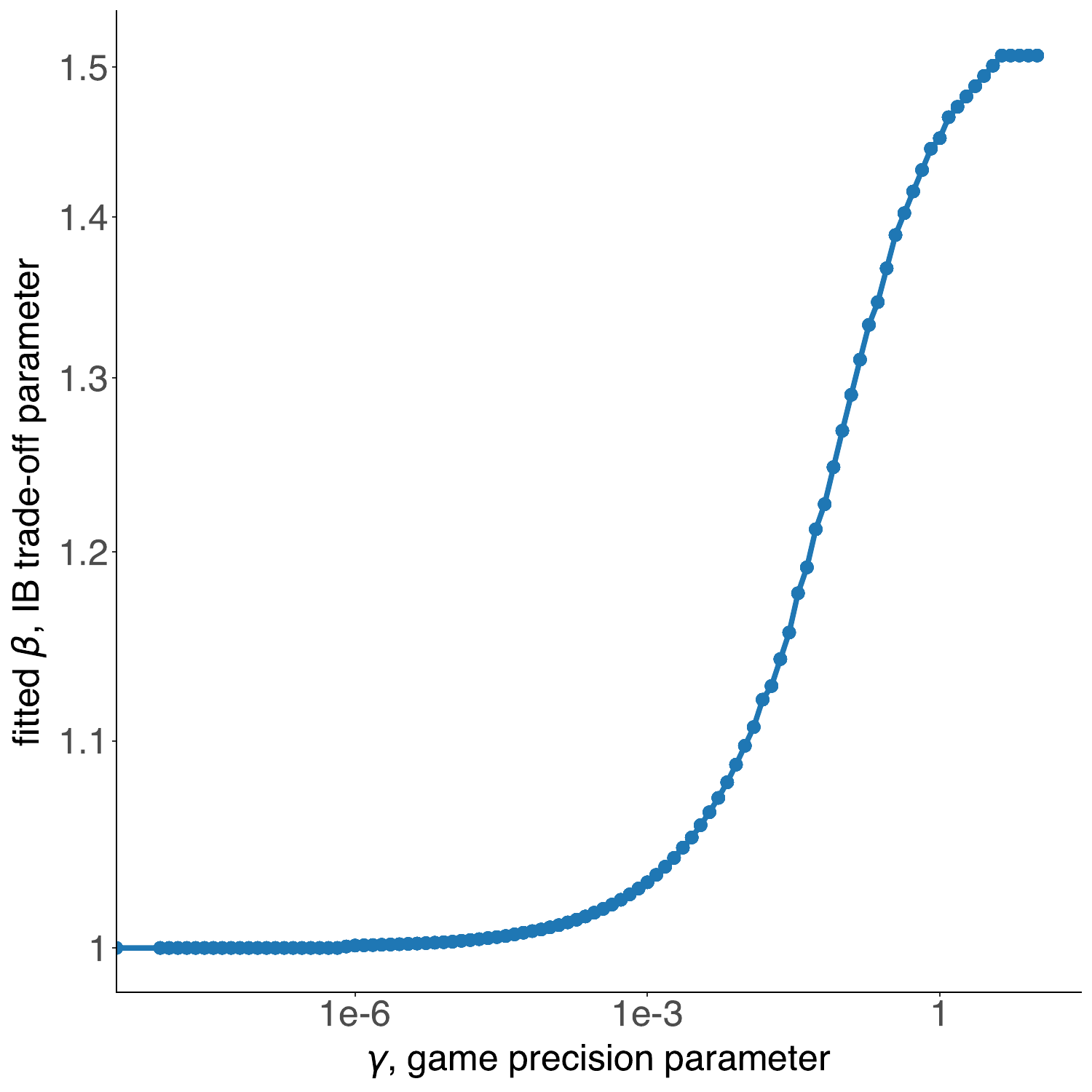}
    \caption{
    Monotonic relationship between $\gamma$ and $\beta$.\\
    }
    \label{fig:gamma_vs_beta}
\end{figure}

\Cref{fig:gamma_vs_beta} shows the monotonic relationship between $\gamma$, the parameter controlling pragmatic standards of precision in each noisy sim-max game, and the IB trade-off parameter values $\beta$ of converged systems, where the latter are fitted by minimum deviation from optimality ($\epsilon$, \Cref{eq:epsilon}). The resulting sigmoidal curve is smooth and highly monotonic, and the Spearman rank correlation between the two variables is very high ($\rho = 0.99, ~p = 0.0$).  While $\gamma$ and $\beta$ are not mathematically related, these results empirically suggest that in the context of noisy imitation of communicative strategies, pragmatic standards of precision are a key potential mechanistic parameter for explaining the range of complexity-accuracy trade-offs to which populations converge.

\section{Baselines}
\label{app:baselines}

We constructed several baselines for the converged systems and for their evolution. To ensure that the efficiency objective is not trivially satisfied, we obtained hypothetical variants of the converged systems by randomly permuting the meanings of each system. We also constructed a dynamical baseline by comparing the trajectories to those resulting from a different evolutionary game dynamic proposed in the literature to model language evolution. We describe these baselines in the following subsection.

\paragraph{Permutations}
For a `static' baseline, we take the converged systems resulting from the imprecise conditional imitation dynamic and permute their meaning-word mappings (i.e., for each system, we permute the rows of their stochastic encoder matrix). We systematically manipulated the randomness of this permutation, and therefore the strength of this baseline, by sampling without replacement from a conditional softmax distribution over the meaning space and varying the temperature parameter, $\tau$. That is, for each converged Sender's encoder, for each $m \in \mathcal{M}$ we swap the distribution $S(w|m)$ with $S(w|m')$, sampling $m' \sim p_{\tau}(m'|m) \propto \exp( \frac{-1}{\tau}(m - m')^{2} )$. Notice that as $\tau \rightarrow \infty$, the permutation is random; as $\tau \rightarrow 0$, the identity permutation is recovered. We used 100 logarithmically spaced values between $1$ and $1000$ for $\tau$.

\paragraph{Finite-population dynamic}
Second, we consider a dynamic baseline. For this, we simulate a dynamic previously studied by \cite{nowakEvolutionLanguage1999} to investigate the evolution of semantic category systems. Informally, this dynamic models the evolution of language within a single, finite population, in which each individual plays with every other individual. Each member of the population plays equally often as a (probabilistic) Sender or Receiver. Fitness is based on communication success, and offspring inherit mappings by sampling their parent’s responses. This finite sampling introduces variation. \cite{nowakEvolutionLanguage1999} showed, in a smaller toy domain, how this dynamic can lead to population-level vocabularies that are evolutionarily stable, yet do not maximize fitness (i.e., correspond to approximate non-strict Nash equilibria). A detailed description of this stochastic process as an algorithm can be found in \Cref{app:nk99}.

We chose this dynamical baseline because it has been independently proposed as a simple model of language evolution, and while related to the replicator dynamic, it differs from our model in several significant ways. Specifically: (1) it is a finite-population model, requiring sampling-based updates to be stochastic rather than deterministic; (2) it models language evolution within a single population by symmetrizing the signaling game, rather than involving two populations; (3) it does not include perceptual uncertainty; and (4) it incorporates mutation, as individuals in each generation acquire their parents' languages through finite sampling. Meanwhile,  \citet{frankeVaguenessImpreciseImitation2018}'s evolutionary dynamic also describes stochastic strategy transmission via imitation, but implements this at the modeling level via deterministic dynamics.

\section{NK99}
\label{app:nk99}

The stochastic process from \cite{nowakEvolutionLanguage1999} is formally described in \Cref{alg:nk}. 

\begin{algorithm}
\caption{Evolutionary Dynamics for Signal-Object Associations}
\begin{algorithmic}[1]
\Require Number of individuals $N$, signals $|\mathcal{W}|$, objects $ |\mathcal{M}| $, generations $T$, number of samples $D$
\State Initialize population: each individual $i$ has random matrices $S_i^0 \in \mathbb{R}^{|\mathcal{W}| \times |\mathcal{M}|}$ and $R_i^0 \in \mathbb{R}^{|\mathcal{M}| \times |\mathcal{W}|}$
\For{$t = 0$ to $T-1$} \Comment{Iterate through generations}
    \State Set total payoff $F_i \gets 0$ for each individual $i$
    \For{each pair of individuals $(i, j)$, $i \neq j$} \Comment{Communication phase}
        \State Compute payoff $F(i, j)$ for $i$ as speaker and $j$ as listener according to \Cref{eq:nk_fitness}.
        \State Update $F_i \gets F_i + F(i, j)$
        \State Update $F_j \gets F_j + F(j, i)$
    \EndFor
    \State Normalize payoffs $F_i$ to compute fitness for each individual:
    $\text{Fitness}_i \gets \frac{F_i}{\sum_{j=1}^{N} F_j}$
    \State Generate offspring proportional to fitness:
    \For{each individual $i$}
        \State Choose parent index $p$ with probability proportional to $\text{Fitness}_p$
        \State Offspring matrices  $S_i^{(t+1)}$ and $R_i^{(t+1)}$ are sampled from parent $p$ mappings:
        \[
        S_{i, mw}^{(t+1)} \gets \text{Sample from } \{S_{p, mw}^{(t)}\}, \quad R_{i, wm}^{(t+1)} \gets \text{Sample from } \{R_{p, wm}^{(t)}\}
        \]
        \State Normalize $S_i^{(t+1)}$ and $R_i^{(t+1)}$ to maintain probabilities
    \EndFor
\EndFor
\State Compute population-level averages of $S$ and $R$ to analyze results
\end{algorithmic}
\label{alg:nk}
\end{algorithm}

The symmetric fitness function from \cite{nowakEvolutionLanguage1999} is:
\begin{equation}
    \label{eq:nk_fitness}
    F(L, L') = \frac{1}{2} \sum_{m} \sum_{w} S(w| m) R'(w | \hat{m}) + S'(w|m) R(w| \hat{m}) 
\end{equation}

There are several things to notice about \Cref{eq:nk_fitness}. First, it assumes a uniform prior over meanings (objects), which is aligned with the setting we have assumed throughout our simulations. Second, it assumes binary payoffs, i.e., a regime in which $\gamma \rightarrow \infty$. Third, it assumes no confusion of meanings for Senders and Receivers, i.e. $p_\alpha(u|m)$ is a delta function with $\alpha \rightarrow \infty$. Interestingly, upon initial exploration of these latter assumptions, we found that (i)  varying $\gamma$ did not systematically change the complexity or accuracy of resulting systems, and (ii) introducing even a small amount of uncertainty $\alpha$ into Senders' and Receivers' perception of meaning resulted in too little selection pressure, i.e. random systems at the information plane origin, which would constitute an uninformative baseline.

To keep simulations finishing under a reasonable amount of time, we ran Algorithm \ref{alg:nk} for $T=100$ generations, assumed $N=20$ individuals, and that each offspring could obtain, for each meaning $m$, $D = 10$ samples from its parents' $S(w|m)$, and likewise $D=10$ samples from its parents' $R(m|w)$ for each word $w$.

\end{document}